\documentclass{article}
\usepackage{ijcai19}

\usepackage{times}
\usepackage{soul}
\usepackage{url}
\usepackage[hidelinks]{hyperref}
\usepackage[utf8]{inputenc}
\usepackage[small]{caption}
\usepackage{graphicx}
\usepackage{amsmath}
\usepackage{booktabs}
\usepackage{algorithm}
\usepackage{algorithmic}
\usepackage{color}
\title{Mask and Infill: Applying Masked Language Model to Sentiment Transfer}

\author{
Xing Wu$^{1,2,3}$\and
Tao Zhang$^{1,2}$\and
Liangjun Zang$^{1}$\and
Jizhong Han$^{1}$\And
Songlin Hu$^{1}$\footnote{Corresponding Author}\\
\affiliations
$^1$Institute of Information Engineering, Chinese Academy of Sciences, Beijing, China\\
$^2$School of Cyber Security, University of Chinese Academy of Sciences, Beijing, China\\
$^3$Baidu Inc., Beijing, China\\
\emails
wuxing03@baidu.com,\{zhangtao,zangliangjun,hanjizhong,husonglin\}@iie.ac.cn
}

\begin{document}

\maketitle

\begin{abstract}

This paper focuses on the task of sentiment transfer on non-parallel text, which modifies sentiment attributes (e.g., positive or negative) of sentences while preserving their attribute-independent content. Due to the limited capability of RNN-based encoder-decoder structure to capture deep and long-range dependencies among words, previous works can hardly generate satisfactory sentences from scratch.
When humans convert the sentiment attribute of a sentence, a simple but effective approach is to only replace the original sentimental tokens in the sentence with target sentimental expressions, instead of building a new sentence from scratch. Such a process is very similar to the task of Text Infilling or Cloze, which could be handled by a deep bidirectional Masked Language Model (e.g. BERT).
So we propose a two step approach ``Mask and Infill". In the \emph{mask} step, we separate style from content by masking the positions of sentimental tokens. In the \emph{infill} step, we retrofit MLM to Attribute Conditional MLM, to infill the masked positions by predicting words or phrases conditioned on the context\footnote{In this paper, \emph{content} and \emph{context} are equivalent, \emph{style}, \emph{attribute} and \emph{label} are equivalent.}and target sentiment.
We evaluate our model on two review datasets with quantitative, qualitative, and human evaluations. Experimental results demonstrate that our models improve state-of-the-art performance.

\end{abstract}

\section{Introduction}

\begin{figure}[!h]
	\begin{centering}
	\includegraphics[width=0.45\textwidth]{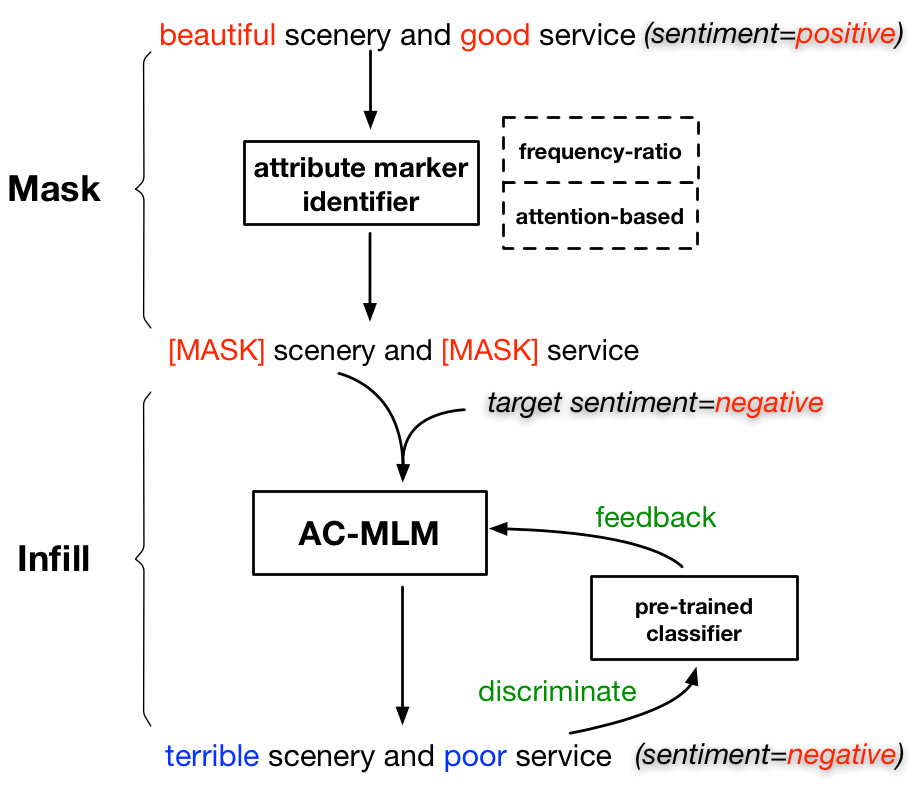}
	\caption{Process of our approach. In the mask stage, we explicitly identify and mask the sentiment tokens in a sentence. In the infill stage, we fill the masked positions with new expressions conditioned on their context and the target sentiment.
} \label{fig1}
	\end{centering}
\end{figure}

The goal of sentiment transfer~\cite{huang2017arbitrary,logeswaran2018content} is to change the sentiment attribute of text while keeping its semantic content unchanged, with broad applications of review sentiment transformation, news rewriting, and so on.
For example, we could convert a positive sentence “I highly recommend this movie" to a negative one “I regret watching this movie".
Lacking supervised parallel data, i.e. pairs of sentences with the same content but different attributes, makes it hard to change the sentiment of text without loss of its semantic content.
Recently, several models have been proposed to learn sentiment transfer from non-parallel text, notably~\cite{shen2017style,DBLP:conf/aaai/FuTPZY18,xu2018unpaired,DBLP:conf/naacl/LiJHL18}.
Some of them~\cite{shen2017style,DBLP:conf/acl/TsvetkovBSP18,DBLP:conf/aaai/FuTPZY18} try to learn the disentangled representation of content and attribute of a sentence in a hidden space, while the others~\cite{xu2018unpaired,DBLP:conf/naacl/LiJHL18} explicitly separate style from content in feature-based ways and encode them into hidden representations respectively.
Afterwards, they utilize a RNN decoder (e.g. LSTM~\cite{hochreiter1997long,sutskever2014sequence,DBLP:conf/emnlp/ChoMGBBSB14}) to generate a new sentence conditioned on the hidden representations of original content and target sentiment attribute.
On the one hand, a shallow RNN encoder structure has limited ability to produce high-quality hidden representation.
On the other hand, usage of RNN decoder restricts prediction ability to a short range, which often fails to produce long realistic sentences.

In this work, we leverage an important observation.
When people convert a sentence with a particular sentiment into one with a different sentiment, we do not have to create a new sentence from scratch. Instead, we simply replace the sentimental tokens of the sentence with other expressions indicative of a different sentiment.
For example, given a negative review “\emph{terrible} scenery and \emph{poor} service", a straightforward positive review is “\emph{beautiful} scenery and \emph{good} service".
Such a process is very similar to the task of text infilling~\cite{DBLP:journals/corr/abs-1901-00158} or Cloze~\cite{taylor1953cloze}.
Given the template \lq\lq \underline{~~~~~~} scenery and \underline{~~~~~~} service", we human feel easy to fill the right words because we have enough prior linguistic knowledge to predict the missing words from their contexts.

We transform the problem of sentiment transfer to the task of text infilling by a pre-trained Masked Language Model (MLM). We utilize a pre-trained deep bidirectional language model (corresponding to human linguistic knowledge) to infill the masked positions.
Our approach comprises two stages: Mask and Infill (Figure~\ref{fig1}).
In the mask stage, we explicitly identify and mask the sentiment tokens in a given sentence.
In the infill stage, we fill the masked positions with new expressions conditioned on their context and the target sentiment.

There exist two feature-based methods for identifying sentiment attribute markers
(i.e. words or phrases that are indicative of a particular sentiment attribute).
Attention-based method~\cite{xu2018unpaired} trains a self-attention\cite{DBLP:journals/corr/BahdanauCB14} sentiment classifier, where the learned attention weights can be used as signals to identify sentiment markers, and the tokens with weights higher than the average are regarded as sentiment attribute markers.
The attention-based method prefers isolated single words to n-gram phrases, which restricts MLM to produce sentences of diverse expressions.
Frequency-ratio method~\cite{DBLP:conf/naacl/LiJHL18} constructs an attribute marker dictionary of n-gram phrases for each sentiment attribute 
and looks up the dictionary to identify sentiment phrases in sentences.
Frequency-ratio method is inclined to mask longer phrases, and its performance heavily relies on the quality of attribute marker dictionaries.
Thus, we explore a simple fused method to utilize the merits of both methods.
Specifically, we combine the attention-based classifier with the frequency-ratio method, filtering out false attribute markers and discovering new single sentiment words.
When the frequency-ratio method fails to identify any attribute marker or recognize too many ones (with insufficient content left), we utilize the attention-based method directly.

MLM predicts the masked tokens only conditioned on their context, considering no attribute information. To fill the masked positions in a sentence with expressions compatible to a particular sentiment, we retrofit MLM to Attribute Conditional Masked Language Model (AC-MLM) by integrating attribute embeddings with the original input for MLM.
Furthermore, we introduce a pre-trained sentiment classifier to constrain AC-MLM, which ensures the generated sentences compatible with the target sentiment.
To deal with the discrete nature of language generation, we utilize soft-sampling~\cite{hu2017toward} to guide the optimization of AC-MLM, where soft-sampling is used to back-propagate gradients through the sampling process by using an approximation of the sampled word vector.

We evaluate our models on two review datasets \emph{Yelp} and \emph{Amazon} by quantitative, qualitative, and human evaluations.
Experimental results show that our method achieves state-of-the-art results on accuracy, BLEU\cite{papineni2002bleu} and human evaluations.

Our contributions are summarized as follows:
\begin{itemize}
\item We propose a two-stage ``Mask and Infill" approach to sentiment transfer task, capable of identifying both simple and complex sentiment markers and producing high-quality sentences.
\item Experimental results show that our approach outperforms most state-of-the-art models in terms of both BLEU and accuracy scores.
\item We retrofit MLM to AC-MLM for labeled sentence generation. To the best of our knowledge, it is the first work to apply a pre-trained masked language model to labeled sentence generation task.
\end{itemize}

\section{Related Work}

\subsection{Non-parallel Style Transfer}

\cite{shen2017style,DBLP:conf/aaai/FuTPZY18,xu2018unpaired,DBLP:conf/naacl/LiJHL18,yang2018unsupervised} are most relevant to our work. \cite{shen2017style} assumed a shared latent content distribution across different text corpora, and leverages refined alignment of latent representations. \cite{DBLP:conf/aaai/FuTPZY18} learned a representation that only contains the content information by multi-task learning and adversarial training.\cite{xu2018unpaired} proposed a cycled reinforcement learning method by collaboration between a neutralization module and an emotionalization module. \cite{DBLP:conf/naacl/LiJHL18} separated attribute and content by deleting attribute markers, and attempted to reconstruct the sentence using the content and the retrieved target attribute markers with an RNN. \cite{yang2018unsupervised} used a target domain language model as the discriminator to guide the generator to produce sentences. Unlike these models, which adopt RNN as encoder and decoder, we utilize pre-trained MLM to capture longer linguistic structure and better language representation.

\subsection{Fine-tuning on Pre-trained Language Model}

Language model pre-training has attracted widespread attention, and fine-tuning on pre-trained language models has proven to be effective in improving many downstream natural language processing tasks. \cite{dai2015semi} improved Sequence Learning with recurrent networks by pre-training on unlabeled data. \cite{radford2018improving} improved the performance largely on many sentence-level tasks from the GLUE benchmarks~\cite{DBLP:conf/emnlp/WangSMHLB18}. \cite{howard2018universal} proposed Universal Language Model Fine-tuning (ULMFiT), which was a general transfer learning method for fine-tuning a language model. \cite{radford2018improving} demonstrated that by generative pre-training language model on unlabeled text from diverse corpora, large gains could be achieved on a diverse range of tasks. By introducing a deep bidirectional masked language model, BERT~\cite{devlin2018bert} obtained new state-of-the-art results on a broad range of tasks. Unlike previous works fine-tuning pre-trained language models to perform discriminative tasks, we aim to apply a pre-trained masked language model on generative task.

\begin{figure}[tp!]
	\begin{centering}
	\includegraphics[width=0.45\textwidth]{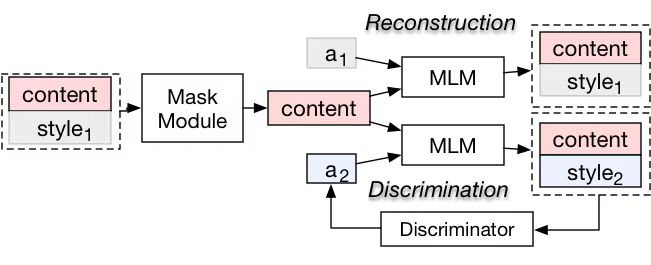}
	\caption{The overall model architecture consists of two modules. The Mask module is the fusion-method introduced in Section 3.2. The Infill module consists of two parts, reconstruction and discrimination, corresponding to Equation 10.} \label{fig2}
	\end{centering}
\end{figure}

\section{Approach}

In this section, we introduce our proposed approach. An overview is presented in Section 3.1. The mask process is introduced in Section 3.2. The details of the infill process are shown in Section 3.3.

\subsection{Overview}

Given a corpus of labeled sentences: ${\cal{D}} = \{(S_1, a_1), . . . ,(S_m, a_m)\}$, where $S_i$ is a sentence and $a_i$ is the corresponding attribute. 
We define ${\cal{A}}$ as the set of all possible attributes, ${\cal{D}}_a = \{S : (S, a) \in {\cal{D}}\}$ as the set of sentences with attribute $a$. 

Our goal is to learn a model that takes as input $(S,\hat{a})$, where $S$ is a sentence with original sentiment attribute and $\hat{a}$ is target sentiment attribute,
outputs a sentence $\hat{S}$ that retains the semantic content of $S$ while exhibiting $\hat{a}$.

The proposed approach consists of two parts: a mask module and an infill module, as shown in Figure~\ref{fig2}. 
The mask module combines the strength of two methods, utilizing a candidate phrase vocabulary to find the potential attribute markers in sentences, then performs mask operation. 
Then the masked sentences are sent into the infill module, which infills masked positions via AC-MLM. During the second stage, for better attribute compatibility of generated sentences, we introduce a pre-trained classifier as the discriminator. Due to the discrete nature of language generation, we draw support from soft-sampling to guide the optimization of our model. The implement details are shown in Algorithm 1.

\subsection{Identify and Mask Attribute Marker}

We first introduce the frequency-ratio method, then the attention-base method. At last, we propose a fusion-method by combing the strength of these two methods.

\subsubsection{Frequency-ratio Method} 

We define for any $a$ in ${\cal{A}}$, let $count(u, {\cal{D}}_a)$ denotes the number of times an n-gram $u$ appear in ${\cal{D}}_a$, we score the salience of $u$ with respect to $a$ by its smoothed frequency-ratio :
\begin{equation}
s_c(u, a) = \frac{count(u, {\cal{D}}_a) + \lambda} {\Big(\sum_{a^{\prime} \in {\cal{A}}, a^{\prime} \neq a} count(u, {\cal{D}}_{a^{\prime}})\Big) + \lambda}  \tag{1}
\end{equation}
where $\lambda$ is the smoothing parameter.
We declare $u$ to be a candidate attribute marker for $a$ if $s_c(u, a)$ is above a specified threshold $\gamma_c$, these candidate attribute markers constitute a candidate attribute marker vocabulary ${{\cal{V}}_a}$.     

\subsubsection{Attention-based Method} 

Each word in a sentence contributes differently to the label or attribute of the whole sentence, we train an attention-based classifier to extract the attribute relative extent.  Given a sequence $\textit{S}$ of N tokens, $<t_1,t_2,...,t_N>$,
we adopt a bidirectional LSTM to encode the sentence and concatenate the forward hidden state and the backward hidden state for each word, and obtain:
\begin{equation}
{\bf{H}} = (h_1, h_2, \cdots, h_N) \tag{2}
\end{equation}
where $N$ is the length of given sentence. The attention mechanism produces a vector of attention weights $\bf{a}$ and a vector of weighted hidden states $\bf{c}$. Finally a softmax layer is followed to transform $\bf{c}$ into probability distribution $\bf{y}$:
\begin{align*}
& {\bf{a}} = softmax({\bf{w}} \cdot tanh({\bf{W}} {\bf{H}}^T)) \tag{3} \\
& \bf{c} = \bf{a} \cdot H \tag{4} \\
& \bf{y} = softmax({\bf{W}}^{\prime} \cdot \bf{c}) \tag{5}
\end{align*} 
where $\bf{w}$, $\bf{W}$, $\bf{W^{'}}$ are projection parameters. After being well-trained, the attention-based classifier can be used to extract attention weights, which capture the sentiment information of each word. For simplicity, following~\cite{xu2018unpaired}, we set the averaged attention value in a sentence as the threshold, words with attention weights higher than the threshold are viewed as attribute markers.

\subsubsection{Fusion-Method} 

However, there are fake attribute markers in the candidate vocabulary\footnote{In the original positive vocabulary of Yelp dataset, ``he did a" scores as high as 26.0, yet it is not a positive attribute marker.}. So we use a pre-trained classifier, i.e. the attention-based classifier, to calculate the probability $p$ of being real for each candidate attribute marker with Equation 5 and update the salience score:
\begin{equation}
s(u, a) = s_c(u, a) * p \tag{6}
\end{equation}

We filter out fake attribute markers with $s(u, a)$ below threshold $\gamma$, obtaining a high-quality attribute marker vocabulary ${\cal{V}}$ for attribute $a$. 

After the vocabulary constructed, we identify and mask as many attribute markers as possible by matching the sentences, because within a sentence style words are usually less than content words. Lastly, if frequency-ratio method identifies no attribute marker or too many attribute markers (with insufficient content left, especially for short sentences), we re-process it with the attention-based method\footnote{Note that we use the attention-based classifier twice: the first time, it is used as discriminator to improve the quality of candidate attribute \textbf{vocabulary}, the second time, it is used to extract attention weights for a \textbf{sentence}.} by identifying the positions with high weights. 

\subsection{Text Infilling with MLM}

\subsubsection{Attribute Conditional Masked Language Model}

Unlike traditional language models which predict tokens based on previously generated tokens, Masked Language Model (MLM) predicts the masked tokens according to their context, rather than reconstructing the entire input.
For the task of sentiment transfer, infilled words or phrases should be consistent to a target attribute.
Therefore, we propose Attribute Conditional Masked Language Model (AC-MLM) to infill masked positions conditioned on both their context and a target label, by adding attribute embeddings into original input embeddings.
Let $\hat{a}\in {\cal{A}}$ be a target attribute, $S = \langle t_1,t_2,...,t_N\rangle$ a sentence of N tokens, $M=\{t_{i_1},...,t_{i_m}\}$ the set of masked tokens, $\overline{S} = S\setminus M$ the set of content tokens.

\subsubsection{Reconstruction} 

Lacking parallel sentiment sentence pairs, so we train the AC-MLM to reconstruct the original sentence $S$ conditioned on the content $\overline{S}$ and original attribute $a$.
The AC-MLM is optimized to minimize the reconstruction error of original masked words:
\begin{equation}
{\cal{L}}_{rec} = -\sum\limits_{a \in {\cal{A}}, t_i \in M} log p( t_i | \overline{S}, a)	\tag{7}
\end{equation}

Well-trained AC-MLM model can take the sentence content $\overline{S}$ and a target attribute $\hat{a}$ as input, and output ${p(\cdot|\hat{a},\overline{S})}$ for each masked position. After all masked positions are infilled, we get the transferred sentence:
\begin{equation}
\hat{S} = \text{AC-MLM}(\overline{S}, \hat{a})	\tag{8}
\end{equation}

Specifically, we use the pre-trained BERT as our MLM, and substitute segmentation embeddings (which is useless when training single sentence) with attribute embeddings.

\subsubsection{Pre-trained Classifier Constraint}

To further improve the transfer accuracy, we introduce a pre-trained sentiment classifier to the AC-MLM, i.e. AC-MLM-SS, ``-SS" indicates with soft-sampling which will be introduced later. 
\paragraph{Discrimination} Given the content $\overline{S}$ and a target attribute $\hat{a}$, when a sentence is successfully transferred, the label predicted by classifier should be consistent to the target attribute. So the AC-MLM-SS is further optimized to minimize the discrimination discrepancy:
\begin{equation}
 {\cal{L}}_{acc} = - log p(\hat{a} | \hat{S})	\tag{9}
\end{equation}
where $\hat{S}$ indicates transferred sentence generated by AC-MLM. \\

By combining Equation 7 and Equation 9 we obtain the objective function of AC-MLM-SS :
\begin{equation}
min_{\theta}{\cal{L}} = {\cal{L}}_{rec} + \eta {\cal{L}}_{acc}	\tag{10}
\end{equation}
where $\theta$ is AC-MLM-SS's parameters and $\eta$ is a balancing parameter.

To propagate gradients from the discriminator through the discrete samples, we adopt soft-sampling.
\paragraph{Soft-sampling} Soft-sampling back-propagates gradients by using an approximation of the sampled word vector. The approximation replaces the sampled token $t_i$ (represented as a one-hot vector) at each step with the probability vector
\begin{equation}
\hat{t}_i \sim softmax({\bf{o}}_t / \tau)	\tag{11}
\end{equation}
which is differentiable w.r.t the AC-MLM’s parameters. The resulting soft generated sentence is fed into the discriminator to measure the fitness to the target attribute. 

\begin{algorithm}
\small
  \caption{Implementation of ``Mask and Infill" approach.}\label{alg1}
  \begin{algorithmic}[1]
  \STATE Pre-train attention-based classifier $\textit{Cls}$ (Eq.2-5)
  \STATE Construct attribute marker vocabulary $\cal{V}$ (Eq.1,6)
  \FOR {every sentence $S$ in $\cal{D}$}
	\STATE Mask attribute markers within ${S}$ by looking up $\cal{V}$, getting $\overline{S}$
	\IF {$\overline{S}$ is too short or $\overline{S}$ is the same as $S$}
	\STATE Re-mask with attention weights calculated by $\textit{Cls}$ (Eq.3)
  	\ENDIF
  \ENDFOR
  \FOR {each iteration i=1,2,...,M}
  	\STATE Sample a masked sentence $\overline{S}$ with attribute $a$
  	\STATE Reconstruct ${S}$ with $\overline{S}$ and $a$, calculating ${\cal{L}}_{rec}$ based (Eq.7)
  	\STATE $\hat{a}$ = the target attribute
	\STATE Construct $\hat{S}$ (Eq.8)
	\STATE calculating ${\cal{L}}_{acc}$ (Eq.9)
  	\STATE Update model parameters $\theta$
  \ENDFOR
  \end{algorithmic}
\end{algorithm}

\section{Experiment}
In this section, we evaluate our method on two review datasets Yelp and Amazon. 
Firstly, we introduce datasets, baseline models, training details, and evaluation metrics. 
Secondly, we compare our approach to several state-of-the-art systems. 
Finally, we present our experimental results and analyze each component in detail.

\subsection{Datasets}
We experiment our methods on two review datasets from~\cite{DBLP:conf/naacl/LiJHL18}: Yelp and Amazon~\cite{he2016ups}, each of which is randomly split into training, validation and testing sets. Examples in YELP are sentences from business reviews on Yelp, and examples in AMAZON are sentences from product reviews on Amazon. Each example is labeled as having either positive or negative sentiment. The statistics of datasets are shown in Table~\ref{tab1}.

\subsection{Baselines}
We compared our methods to existing relevant models: CrossAligned~\cite{shen2017style}, StyleEmbedding and MultiDecoder~\cite{DBLP:conf/aaai/FuTPZY18}, CycledReinforce~\cite{xu2018unpaired}\footnote{\url{https://github.com/lancopku/Unpaired-Sentiment-Translation}}, TemplateBased, RetrievalOnly, DeleteOnly and DeleteAndRetrieval~\cite{DBLP:conf/naacl/LiJHL18}, LM+Classifer~\cite{yang2018unsupervised}.

\subsection{Experiment Details}

\subsubsection{Mask step} 

For the frequency-ratio method, we consider n-gram up to 4-gram and set the smoothing parameter $\lambda$ to 1. Other hyperparameters are following~\cite{DBLP:conf/naacl/LiJHL18}. For the attention-based method, we train the attention-based classifier for 10 epochs with accuracy 97\% for Yelp and 82\% for Amazon. For the fusion-method, we consider a sentence with content less than 5 tokens as insufficient content, and re-process it with the attention-based method. 

\subsubsection{Infill step} 

We use pre-trained BERT$_{base}$ as MLM and substitute segment embedding layer with attribute embedding layer. The input size is kept compatible with original BERT and relevant hyperparameters can be found in~\cite{devlin2018bert}. The pre-trained discriminator is a CNN-based classifier~\cite{DBLP:conf/emnlp/Kim14} with convolutional filters of size {3, 4, 5} and use WordPiece embeddings. The hyperparameters in Equation 10 and 11 are selected by a grid-search method using the validation set. We fine-tune BERT to AC-MLM for 10 epochs, and further train 6 epochs to apply discriminator constraint.

\begin{table}[tb]
\small
\centering
\begin{tabular}{ccccc}
\toprule
Dataset &Attributes &Train &Dev &Test \\
\midrule
YELP& Positive &270K &2000 &500 \\
 & Negative&180K&2000 &500 \\
\midrule
AMAZON& Positive &277K&985&500 \\
 & Negative &278K&1015&500 \\
\bottomrule
\end{tabular}
\caption{Dataset statistics.}\label{tab1}
\end{table}

\subsection{Evaluation}

\begin{table}[!b]
\small
\centering
\begin{tabular}{lrrrr}
\toprule
& \multicolumn{2}{c}{YELP}& \multicolumn{2}{c}{AMAZON} \\
& ACC (\%)& BLEU& ACC(\%)& BLEU\\
\midrule
CrossAligned& 73.7& 3.1& 74.1& 0.4\\
StyleEmbedding& 8.7& 11.8& 43.3& 10.0\\
MultiDecoder& 47.6& 7.1& 68.3& 5.0\\
CycledReinforce & 85.2& 9.9& 77.3& 0.1\\
TemplateBased& 81.7& 11.8& 68.7& 27.1\\
RetrievalOnly& 95.4& 0.4& 70.3& 0.9\\
DeleteOnly& 85.7& 7.5& 45.6& 24.6\\
DeleteAndRetrieval& 88.7& 8.4& 48.0& 22.8\\
\midrule
w/frequency-ratio\\
AC-MLM& 55.0& 12.7& 28.7& 31.0\\
AC-MLM-SS& 90.5& 11.6& 75.7& 26.0 \\
\midrule
w/attention-based\\
AC-MLM& 41.5& \textbf{15.9}& 31.2& \textbf{32.1}\\
AC-MLM-SS& 97.3& 14.1& 75.9& 28.5 \\
\midrule
w/fusion-method\\
AC-MLM& 43.5& 15.3& 42.9& 30.7\\
AC-MLM-SS& \textbf{97.3}& 14.4& \textbf{84.5}& 28.5\\
\bottomrule
\end{tabular}
\caption{Automatic evaluation performed by tools from \protect\cite{DBLP:conf/naacl/LiJHL18}\protect\footnotemark. ``ACC'' indicates accuracy, ``BLEU" measures content similarity between the outputs and the human references. ``AC-MLM", represents attribute conditional masked language model. ``w/" represents ``with".``-SS" represents with soft-sampling.} \label{tab2}
\end{table}
\footnotetext{https://github.com/lijuncen/Sentiment-and-Style-Transfer}

\begin{table}[!tb]
\centering
\small
\begin{tabular}{lrr}
\toprule
& ACC (\%)& BLEU\\
\midrule
Yang's results\\
LM& 85.4& 13.4\\
LM+Classifier& 90.0& \textbf{22.3}\\
\midrule
w/frequency-ratio\\
AC-MLM&  58.0&  18.7\\
AC-MLM-SS&  93.7&  17.5\\\midrule
w/attention-based\\
AC-MLM& 40.0 & 21.8\\
AC-MLM-SS& \textbf{98.5} & 20.5 \\\midrule
w/fusion-method\\
AC-MLM&  44.1&  21.3\\
AC-MLM-SS& 97.3& 20.7 \\
\bottomrule
\end{tabular}
\caption{Automatic evaluation on Yelp dataset performed by tools from \protect\cite{yang2018unsupervised}\protect\footnotemark.} \label{tab3}
\end{table}
\footnotetext{\cite{yang2018unsupervised} only evaluated their models on Yelp. For fair comparison, we only evaluate our models on Yelp too.}

\subsubsection{Automatic Evaluation}

We compute automatic evaluation metrics by employing automatic evaluation tools from~\cite{DBLP:conf/naacl/LiJHL18} and~\cite{yang2018unsupervised}. Accuracy score assesses whether outputs have the desired attribute. BLEU score is computed between the outputs and the human references. A high BLEU score primarily indicates that the system can correctly preserve content by retaining the same words from the source sentence as the reference. 

Table~\ref{tab2} shows the performance obtained with~\cite{DBLP:conf/naacl/LiJHL18}’s tools. Our base model AC-MLM achieves all the best BLEU, but performs poorly in accuracy. After applying the discriminator constraint to apply constraint (i.e. AC-MLM-SS), the accuracy improves significantly, though with the slight decline of BLEU. Previous models are hard to perform well on accuracy and BLEU simultaneously. Among our models, AC-MLM-SS using fusion-methods in \emph{mask} step achieves the most satisfactory performance considering on both accuracy and BLEU.
Table~\ref{tab3} shows the performance on~\cite{yang2018unsupervised}’s tools. Compared to Yang’s best model LM+Classifier, we perform better on accuracy, but slightly lower on BLEU.

\begin{table*}[!t]
\centering
\begin{tabular}{ll}
\toprule
\multicolumn{2}{c}{From negative to positive (YELP)}\\
\midrule
Source & it 's \textcolor{red}{not much like} an actual irish pub , which is \textcolor{red}{depressing} .\\
Human & It's \textcolor{blue}{like} an actual irish pub .\\
CrossAligned & it 's \textcolor{blue}{not good} for a clean and inviting , i textcolor{blue}{love} food .\\
StyleEmbedding & it 's not much like an neat of vegetarian - but tiny crust .\\
MultiDecoder & it 's not much like an vegetarian bagel ... much is food .\\
CycledReinforce &it 's not much like an actual irish pub , \textcolor{blue}{excellent sweet sweet} !\\
TemplateBased & it 's not much like an actual irish pub , which is most \textcolor{blue}{authentic} .\\
RetrievalOnly & i \textcolor{blue}{like} their food , i \textcolor{blue}{like} their prices and i \textcolor{blue}{like} their service .\\
DeleteOnly & it 's not like much an actual irish pub , which is very \textcolor{blue}{fun} .\\
DeleteAndRetrieval & it 's not much like an actual irish pub , which is \textcolor{blue}{my favorite }. \\
AC-MLM & it 's not much like an actual irish pub , which is quite \textcolor{blue}{nice} .\\
AC-MLM-SS & it 's \textcolor{blue}{pretty much like} an actual irish pub , which is very \textcolor{blue}{fantastic}.\\
\midrule
\end{tabular}
\caption{Example outputs on YELP. ``Human" line is a human annotated sentence. Original negative attribute markers are colored in red, transferred positive ones are colored in blue.}\label{tab5}
\end{table*}

\begin{table}
\small
\centering
\begin{tabular}{lrrrrrr}
\toprule
& \multicolumn{3}{c}{YELP}& \multicolumn{3}{c}{AMAZON} \\
& Gra& Con& Att& Gra& Con& Att\\
\midrule
DeleteAndRetrieval& 3.4& 3.5& 3.6& 3.5& 3.2& 3.3\\
\midrule
w/frequency-ratio\\
AC-MLM-SS& 3.9& 3.2& 4.2& 3.8& 3.6& 3.7\\
\midrule
w/attention-based\\
AC-MLM-SS& 4.0& 3.8& \textbf{4.4}& 3.9& 3.7& 3.7\\
\midrule
w/fusion-method\\
AC-MLM-SS& \textbf{4.2}& \textbf{4.0}& \textbf{4.4}& \textbf{4.1}& \textbf{4.0}& \textbf{4.0}\\
\bottomrule
\end{tabular}
\caption{Human evaluation results on two datasets. We show average human ratings for grammaticality (Gra), content preservation (Con), target attribute match (Att).}\label{tab4}
\end{table}

\subsubsection{Human Evaluation}

We hired three annotators (not authors) to rate outputs for our models and \cite{DBLP:conf/naacl/LiJHL18}'s best model (DeleteAndRetrieval). We adopt three criteria range from 1 to 5 (1 is very bad and 5 is very good): grammaticality, similarity to the target attribute, and preservation of the source content. For each dataset, we randomly sample 200 examples for each target attribute. Table~\ref{tab4} shows the human evaluation results. Among them, our AC-MLM-SS with the fusion-method achieves best results. 
Pre-trained MLM infills the masked positions considering the context from both directions, being able to meet grammatical requirements better. Benefiting from explicitly separating content and style, our sentences' structures are kept and we perform better on the preservation of the source content. Then by introducing the pre-trained discriminator, MLM infills more accurate attribute words.

\section{Analysis}

\begin{figure}[!tb]
	\begin{centering}
	\includegraphics[width=0.5\textwidth]{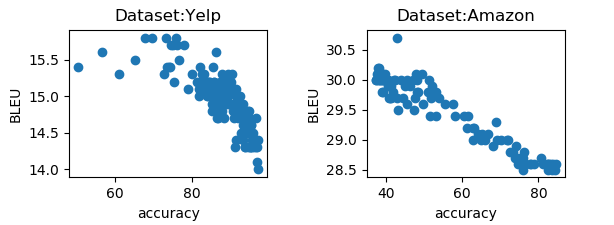}
	\caption{The trend of BLEU with the increase of accuracy.} \label{fig3}
	\end{centering}
\end{figure}

\paragraph{Trade-off between Content and Attribute}
Our loss function in Equation 10 consists of two parts: reconstruction loss and discriminative loss. Reconstruction loss guides MLM to infill contextual-compatible words. Compared to weak label-constraint introduced by attribute embeddings, discriminative loss from classifier improves accuracy largely by encouraging more attribute-compatible words, which decreases the diversity. The two losses are adversarial and cooperate to balance the accuracy and BLEU of transferred sentences. We plot the trade-off curve of the two indicators in Figure~\ref{fig3}. We can modify the hyperparameter $\eta$ in Equation 10 to control the trend. 

\paragraph{Comparing Three Mask Methods}
In general, fusion-method performs best on automatic evaluation. On the high-quality Yelp dataset, fusion-base method gains slight improvement and performs closely to the attention-based method. But on the Amazon dataset, not as good as Yelp, the fusion-method gains large improvement by combining the merits of the attention-based and frequency-ratio methods.

\paragraph{Benefits of Masked Language Model}
Compared to the models using LSTM decoder in Table~\ref{tab2}, our models highly improve the BLEU on both datasets. Compared to~\cite{yang2018unsupervised} in table~\ref{tab3}, which introduces the target domain language model as the discriminator, we also achieve comparable performance. But training a target domain language model from scratch is very time consuming and in need of large corpus. We directly fine-tune on the pre-trained BERT.

\paragraph{Connection to PG-Net}
Our models can be viewed as a collaboration between copying and generating, which is similar to the hybrid pointer-generator network~\cite{DBLP:conf/acl/SeeLM17} for text summarization task. Copying aids accurate reproduction of content information, while retaining the ability to produce stylish words through the generator. Unlike PGNet which copies words from the source text via pointing, our approach copies words from the source via masking. 

\section{Conclusions and Future Work}
In this paper, we focus on non-parallel sentiment transfer task and propose a two-stage ``mask and infill" approach that enables training in the absence of parallel training data. Experimental results on two review datasets show that our approach outperforms the state-of-the-art systems, both in transfer accuracy and semantic preservation. For future work, we would like to explore a fine-grained version (more than two sentiments) of sentiment transfer, and explore how to apply the masked language model to other tasks of natural language generation beyond style transfer.

\section*{Acknowledgments}
This research is supported in part by  the National Key Research and Development Program of China under Grant 2018YFC0806900 and 2017YFB1010000, the Beijing Municipal Science and Technology Project under Grant Z181100002718004.

\newpage
\bibliographystyle{named}
\bibliography{ijcai19}

\end{document}